\documentclass{article}

\usepackage[preprint, nonatbib]{neurips_2020}

\usepackage[utf8]{inputenc} 
\usepackage[T1]{fontenc}    
\usepackage{hyperref}       
\usepackage{url}            
\usepackage{booktabs}       
\usepackage{amsfonts}       
\usepackage{nicefrac}       
\usepackage{microtype}      

\usepackage{caption}
\usepackage{subcaption}
\usepackage{amsthm}

\usepackage{color}
\usepackage{bbding}
\usepackage{bm}
\usepackage{float}
\usepackage{lipsum}
\usepackage{amsmath}
\usepackage{amssymb}
\usepackage{graphicx}

\DeclareMathOperator*{\argmax}{argmax-k} 

\title{Pruning Convolutional Filters using Batch Bridgeout}
\author{%
  Najeeb Khan\\
  Department of Computer Science\\
  University of Saskatchewan, SK, Canada \\
  \texttt{najeeb.khan@usask.ca} \\
  \And
  Ian Stavness \\
  Department of Computer Science\\
  University of Saskatchewan, SK, Canada \\
  \texttt{ian.stavness@usask.ca} \\
}

\begin{document}
\maketitle
\begin{abstract}
   State-of-the-art computer vision models are rapidly increasing in capacity, where the number of parameters far exceed the number required to fit the training set. This results in better optimization and generalization performance. However, the huge size of contemporary models results in large inference costs and limits their use on resource-limited devices. In order to reduce inference costs, convolutional filters in trained neural networks could be pruned to reduce the run-time memory and computational requirements during inference. However, severe post-training pruning results in degraded performance if the training algorithm results in dense weight vectors. We propose the use of Batch Bridgeout, a sparsity inducing stochastic regularization scheme, to train neural networks so that they could be pruned efficiently with minimal degradation in performance. We evaluate the proposed method on common computer vision models VGGNet, ResNet and Wide-ResNet on the CIFAR image classification task. For all the networks, experimental results show that Batch Bridgeout trained networks achieve higher accuracy across a wide range of pruning intensities compared to Dropout and weight decay regularization.
\end{abstract}



\section{Introduction}
The combination of larger GPUs and more effective regularization techniques, such as Dropout~\cite{srivastava2014dropout} and BatchNorm~\cite{ioffe2015batch}, has enabled deep learning practitioners to train larger models with better generalization performance compared to smaller models. From LeNet in 1990s with less than 1 million parameters, to mixture of expert Deep Neural Networks (DNNs) consisting of up to 137 billion parameters~\cite{ShazeerMMDLHD17}, task performance has improved significantly in many challenging domains \cite{xie2019self,hestness2017deep}.  

The increase in size and performance comes at the cost of huge computational requirements both for training and inference. Beside computational requirements, contemporary DNN based computer vision models have a huge run-time memory footprint due to their large number of parameters. Deep neural networks are increasingly deployed to resource-limited devices such as smart phones and internet-of-things devices, where these large models would not fit within the memory constraints of the device.

Techniques to reduce the inference cost of DNNs are needed to make DNNs useful for deployment to edge devices and scalable in server applications. Several approaches have been proposed for efficient inference in DNNs including: sharing weights between different neural units or layers~\cite{aich2020multi,ullrich2017soft,inan2016tying}, quantizing the weights of the model to fewer bits in order to lower the spatial and temporal cost~\cite{gysel2018ristretto}, decomposition of large weight matrices into smaller tensors for fast processing~\cite{novikov2015tensorizing}, distilling the knowledge of a large model into a smaller one~\cite{hinton2015distilling}, and pruning.  

Pruning is the most commonly used technique to reduce the inference cost of a trained neural network and is inspired by the decline of synaptic density in the human cerebral cortex with age~\cite{huttenlocher1979synaptic}. 
In pruning, the elements of a trained network are ranked according to some importance criteria and the least important elements are removed from the network, resulting in a smaller network with lower inference cost. Unstructured pruning removes individual weights from the model resulting in sparse matrices with the same dimensions as the original model, requiring sparse computational techniques to save inference cost~\cite{gale2020sparse}. Structured pruning, on the other hand, removes filters from the convolutional layer reducing the dimension of the matrix multiplication and directly resulting in a smaller number of operations and runtime memory.

The simplest and most common importance criteria for removing filters is the magnitude of the filters. Therefore, several regularization approaches have been proposed to train CNNs such that most weights have smaller values at the end of the training to facilitate pruning \cite{bui2019ell_0,louizos2018learning}. In order to train neural networks that are robust to post-hoc pruning, stochastic regularization techniques provide the advantage that they make the network rely less on any individual weights in the network during training. One such stochastic regularization technique is Bridgeout~\cite{khan2018bridgeout}, proposed for fully connected neural networks. Bridgeout, along with being stochastic, is also proven theoretically to induce sparsity in the model weights.

In this work, we propose a simple and computationally less expensive variant of Bridgeout called Batch Bridgeout. Batch Bridgeout is applicable to both fully connected and convolutional layers. Batch Bridgeout is significantly faster compared to Bridgeout and requires less GPU memory. Importantly, Batch Bridgeout can be easily implemented without changing the optimized GPU based kernels such as cuDNN \cite{chetlur2014cudnn}. Unlike Bridgeout that uses a different set of weights per \textit{example}, Batch Bridgeout uses a single set of weights per \textit{mini-batch of examples}. We show that Batch Bridgeout results in sparse filter weights in CNNs similar to Bridgeout inducing sparsity into fully connected layers.  

Given its stochastic nature, sparsity inducing characteristics, and ease of use, we propose the use of Batch Bridgeout for pruning convolutional filters in CNNs. The stochastic nature of Batch Bridgeout regularization makes the CNN robust to pruning since the network does not rely on any single filter, whereas the sparse nature of the regularization results in networks that distill their knowledge in a smaller set of weights making the network naturally appropriate for pruning.

We train contemporary computer vision models such as VGG-16~\cite{simonyan2014very}, a very small ResNet~\cite{he2016deep} and a large Wide-ResNet~\cite{zagoruyko2016wide}, with Batch Bridgeout targeted to the largest magnitude filter weights, on the CIFAR image classification task. We perform \textit{one-shot structured pruning} of the filters in the networks and show that Batch Bridgeout results in the least degradation compared to Targeted Dropout for all the networks.

The contributions of this work include:
\begin{enumerate}
\item a novel stochastic regularization method, Batch Bridgeout, that can be used to induce sparsity into CNNs while also being easy to implement and computationally less expensive;
\item the novel application of one-shot filter pruning with sparsity inducing regularization; and,
\item the evaluation of structured pruning with Batch Bridgeout across a range of DNN models.
\end{enumerate}

\section{Background and Related Work}
\label{sec:bg}
To make state-of-the-art computer vision models more practical to deploy, many pruning techniques have been proposed. This section provides a taxonomy of pruning techniques followed by a description of the works related to the main contributions of this paper.

\subsection{Classification of Pruning Techniques}
We broadly classify pruning techniques for lower inference cost based on the elements pruned, the number of train-prune iterations and the criteria used for pruning decisions as follows.

\subsubsection{Unstructured vs. Structured Pruning}
Unstructured pruning removes \textit{individual weights} from the weight matrices of convolutional filters or the fully connected units~\cite{han2015learning}. Unstructured pruning results in higher compression rates for the same task performance due to the flexibility of fine grained selection of which weights should be eliminated. Unstructured pruning results in sparse weight matrices with the same dimensions as the unpruned ones. Thus, specialized sparse matrix multiplication techniques are needed to exploit the sparsity for faster inference~\cite{gale2020sparse}.

Structured pruning, on the other hand, removes model weights in groups corresponding to a neural unit or convolutional filter~\cite{li2016pruning}. Structured pruning corresponds to removing an entire row or column from the weight matrix. This results in reduced dimension weight matrices directly reducing the number of operations and runtime memory required for inference without any additional overhead or specialized techniques. However, imposing such structure during pruning generally results in higher degradation in task performance.

\subsubsection{Iterative vs. One-shot Pruning}
Iterative pruning trains a network to convergence, computes the importance of each element in the network and removes a small number of least important elements. This is followed by retraining to recover from any loss in task performance due to pruning. This process is repeated until the desired network size and task performance trade off is achieved~\cite{han2015learning}.

Conversely, in one-shot pruning, the training algorithm is modified in such a way that at the end of training most of the weights (or neurons) are zero. Thus, these weights or neurons could be removed at the end of training without any additional steps nor substantial loss of accuracy. The cost function of DNNs over the weight space has many local minima. One-shot pruning from scratch schemes employ some form of regularization to prefer DNNs with sparse weights compared to dense ones with equivalent cost. This sparsity inducing characteristic of the regularization helps in retaining the network performance when the network is pruned. Several deterministic sparse regularization techniques have been proposed for one-shot pruning of neural networks \cite{yoon2017combined, wen2017learning}.

\subsubsection{Sensitivity vs. Magnitude based Pruning}
When pruning a model, it can be helpful to quantify the importance of weights or neurons in the network. This importance metric is intended to quantify the degradation in performance brought about by removing the weight or neuron. 

In sensitivity based pruning methods, the sensitivity of the cost function with respect to weights or neurons is directly approximated using the derivatives of the cost function with respect to each network element. Optimal Brain Damage (OBD)~\cite{lecun1989optimal} is a technique that uses a second order Taylor approximation of the cost function with respect to individual weights. This approximation requires the computation of the Hessian of the cost function with respect to individual weights. The Hessian is then used as a proxy for the sensitivity of the cost function with respect to the weights.

Instead of using Hessian-based methods, the importance of a network element could be approximated using the magnitude of the network element. A higher magnitude of an element means that element contributes more to the output of the network compared to a low-magnitude element~\cite{han2015learning}. In a large scale empirical study, Gale et al.~\cite{gale2019state} found that simple magnitude-based importance criteria performs better than other complex criteria for pruning deep neural networks.  

In this work we are concerned with one-shot, structured magnitude-based pruning of filters in convolutional neural networks.

\subsection{Sparse Regularization for Pruning}
The purpose of magnitude-based pruning techniques is to remove small magnitude weights or neurons from the network with as little performance drop as possible. It is logical then to train neural networks, from scratch, in such a way that most of the weights or neurons are close to zero except a few weights that are critical to the performance of the network. The cost function of DNNs often exhibit a large number of local minima \cite{Safran2016}. It is, therefore, probable that two local minima are almost equal in magnitude but correspond to very different configurations of the parameters, for example, one could belong to a highly dense DNN, whereas the other could belong to a very sparse, and thus compact, DNN. To this end several deterministic regularization techniques have been explored to train DNNs so that the sparse configurations of the parameters are selected, which in turn, can aid in model pruning.

Alvarez and Salzmann~\cite{alvarez2016learning} have used group sparsity to learn the number of neurons per layer in a deep neural network. They added an $L_{1,2}$ penalty term to the cost function in order to force groups of parameters belonging to a single neuron close to zero.
Evaluating $L_{1,2}$ on Imagenet, the authors reported better compression performance compared to an $L_1$ penalty. The grouped sparsity removes complete neurons and thus this technique is a structured pruning method.

Scardapane et al.~\cite{scardapane2017group} augmented the $L_{2,1}$ penalty with an $L_1$ penalty to avail additional compression for fully connected neural networks using both structured ($L_{1,2}$) and unstructured ($L_{1}$) pruning at the same time. Yoon and Huang~\cite{yoon2017combined} combined the group sparsity ($L_{2,1}$ penalty) with exclusive sparsity ($L_{1,2}$ penalty) to achieve more compact fully connected and convolutional neural networks.

In the previous methods, $L_1$-like penalties were used to drive weights towards zero during training. While the $L_1$ penalty prefers sparser weights it does not make the weights exactly zero. A more representative penalty for non-zero weights is the $L_0$ norm, which is defined as the number of non-zero elements in a vector. However, training with an $L_0$ norm regularizer based on gradient descent is not feasible because the $L_0$ norm is not differentiable. The $L_1$ norm, as described previously, is used as a convex relaxation to the $L_0$ norm.  A few algorithms for training neural networks with an $L_0$ penalty have been proposed in the literature recently. Louizos et al.~\cite{louizos2018learning}  proposed a method to minimize the expected $L_0$ norm of the weights during training followed by unstructured pruning.  Bui et al.~\cite{bui2019ell_0} derived a method for minimizing the group $L_0$ norm in order to perform structured pruning.

Stochastic regularization such as Dropout has been utilized to aid in pruning performance as well. Variational Dropout (VD) uses an individual dropout rate for each parameter during training. Parameters with large Dropout probability at the end of training could thus be discarded. Molchanov \cite{molchanov2017variational} reported good compression performance of VD on LeNet and VGG architectures.

Dropout regularization promotes robustness to the loss of individual neurons during training. During each forward pass Dropout randomly prunes neurons. If the least useful neurons are known a priori, based on some criteria such as magnitude, Dropout could be applied only to these neurons. Such targeted application of Dropout will enable the network to be robust to post-hoc pruning. This idea was proposed by Gomez et al. \cite{gomez2019learning} naming it Target Dropout. Gomez et al. showed that Targeted Dropout resulted in superior performance to other complex pruning strategies. We chose Targeted Dropout as the primary baseline for comparison in our study due to its state-of-the-art performance.

While Dropout zeros out neurons during training and promotes robustness, in expectation, it does not minimize a cost function that promotes sparsity. Therefore, our proposition is that an alternative stochastic regularization scheme, that explicitly induces sparsity, should better retain model accuracy after pruning.

\section{Proposed Method}
\label{sec:method}
In previous work it was observed that deterministic \textit{sparse} regularization techniques ($L_1, L_{1,2}, L_{0}$) promoting sparsity have been utilized to improve pruning performance. In addition, \textit{stochastic} methods such as Dropout have been shown to promote robustness to pruning. Therefore, we expect that the \textit{sparse stochastic} regularizers such as Batch Bridgeout could be used to combine the benefits of both sparsity and robustness for obtaining efficient and compact DNNs through pruning.

\subsection{Batch Bridgeout}
Batch Bridgeout generates a Bernoulli random mask $M$ for each \textit{mini-batch} in the training set, for each hidden-layer in the network. During training Batch Bridgeout applies the following perturbation to the weights of the $l^{th}$ layer $W^{l}$
\begin{equation}
	\widetilde{W}_{ij}^l = \begin{cases} 
		W_{ij}^l - |W_{ij}^l|^{\frac{q}{2}} & \text{if}\;  M_{ij} = 0 , \\
		W_{ij}^l + |W_{ij}^l|^{\frac{q}{2}}\big(\frac{1-p}{p}\big) & \text{if}\;  M_{ij} =1, \\
	\end{cases}
	\label{eq:boPerturb}
\end{equation}
where $p$ is the probability of the Bernoulli mask determining the strength of regularization and $q$ is the norm of the penalty determining the sparsity of the weights. The output of the $l$-th layer is then calculated as
\begin{align}
	\boldsymbol{\nu}^l &= \widetilde{\boldsymbol{W}}^l \boldsymbol{a}^{l-1} + \boldsymbol{b}^l,\\
	\boldsymbol{a}^l &= \sigma\big( \boldsymbol{\nu}^l \big),
\end{align}
where $\boldsymbol{a}^{l-1}$ and $\boldsymbol{a}^{l}$ are the activations of the previous and current layer, respectively. $\boldsymbol{b}^l$ is the bias vector and  $\sigma$ is a non-linear activation function such as sigmoid or ReLU \cite{glorot2011deep}.

The perturbation in Equation~\ref{eq:boPerturb} is equivalent to an $L_q$ penalty on the weights of linear models \cite{khan2018bridgeout}. $L_q$ weight penalties for $q<2$ results in sparse weight matrices. Thus, Batch Bridgeout with $q<2$ enables us to obtain equivalent results as applying sparsity penalties on the weights . At the same time it is a stochastic technique. Therefore, it has the advantage of making the DNNs not rely on any individual weights and hence making the networks robust to any post-hoc pruning of the weights.

Generating a new set of perturbed weights per layer in a deep neural network for each example in a large dataset is prohibitively expensive. Per example perturbation also requires customization of the common GPU based implementations of CNNs such as cuDNN. Batch Bridgeout uses a single set of perturbed weights per mini-batch of examples. We show that using a single set of perturbed weights per layer per mini-batch makes Bridgeout only fractionally more expensive compared to Dropout while still keeping the sparsity inducing properties.

\subsection{Filter Pruning}
Once the network is trained with Batch Bridgeout, many of the weights in the filters will be close to zero due to the sparsity inducing property of the regularization. We use the $L_2$ norm of the filter weights as a surrogate for the importance of the filter to the network. In each layer we keep only the $k$ filters with the largest $L_2$ norms
\begin{equation}
    \mathbf{W}^* = \argmax_j  \|\mathbf{W}_{:,j}\|,
\end{equation}
where $\boldsymbol{W}$ is the matrix containing all the filters in the layer and $\boldsymbol{W}^*$ contains the $k$ most important filters, with the largest $L_2$ norm, we are interested in after pruning.

\begin{figure}
    \centering
    \includegraphics[width=0.9\linewidth]{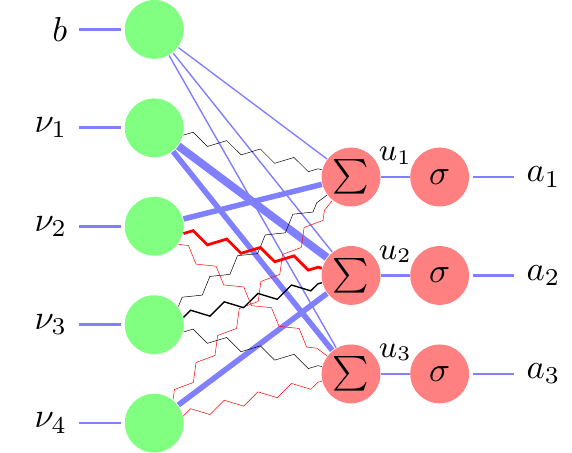}
    \caption{A fully connected layer with four inputs and three outputs. The width of the edges represents the magnitudes of the weights. The squiggly edges represent the Batch Bridgeout perturbation in Equation~\ref{eq:boPerturb}. During training the 8 lowest magnitude weights out of 12 ($\gamma = 0.67$) are targeted by Batch Bridgeout. As training updates the weights, a potentially different set of weights are targeted by Batch Bridgeout. The process is the same for convolutional layers.}
    \label{fig:pruning_t_batch_bo}
\end{figure}

Batch Bridgeout could be applied to all the weights of a DNN layer. However, for the purposes of pruning, our goal is to find the optimal set of  filter weights $\boldsymbol{W}^*$ with respect to the cost function such that the number of non-zero filters is bounded $|\boldsymbol{W}_{:,j}^*|<k$. That is, we want to keep only the $k$ most important filters at the end of training. Rather than dropping weights deterministically at the end of training, it is more reasonable to target the application of Batch Bridgeout to only the less important weights as is done in Targeted Dropout~\cite{gomez2019learning}. If during training some less important weights become significant, Batch Bridgeout is not applied to them. Thus, Targeted Batch Bridgeout helps in distilling the weights of the $k$ most important filters during training. Targeted Batch Bridgeout is illustrated in Figure~\ref{fig:pruning_t_batch_bo} for the case of fully connected layers where it is applied only to a fraction $\gamma$ of the weights in the layer at each iteration of the training.

\section{Experimental Results}
\label{sec:exp}
In this section we describe the experiments that evaluate the computational cost and sparsity induction and performance of Batch Bridgeout. For evaluating the pruning performance, we benchmark Batch Bridgeout against the recently proposed Targeted Dropout \cite{gomez2019learning} technique which has been shown to perform better than other deterministic and stochastic techniques. To asses whether the pruning results generalize to different architectures of different sizes we conducted experiments with three architectures of different sizes: VGG, a very small ResNet and a full sized Wide-ResNet.

\subsection{Computational Cost}
Batch Bridgeout is computationally efficient compared to Bridgeout. To evaluate the computational cost of Batch Bridgeout against Bridgeout, we train a fully-connected autoencoder with two hidden layer sizes on MNIST with a batch size of $128$ both on a Nvidia GTX 1080 GPU.  Table~\ref{tab:computeCost} shows the average execution time per epoch. As can be seen in the table, Batch Bridgeout and Dropout incur similar cost whereas Bridgeout is an order of magnitude slower on the same hardware. Doubling the number of hidden units in the autoencoder results in a two-fold increase in computational time for Bridgeout, whereas Batch Bridgeout's execution time stays almost constant due to GPU parallelism.

\begin{table}
    \centering
    \caption{Average execution time in seconds per epoch for an autoencoder with different hidden layer sizes and regularization methods.}
    \begin{tabular}{|c|c| c |c |} 
         \hline
         Units &  Weight Dropout & Batch Bridgeout & Bridgeout \\ [0.5ex]
         \hline
         1024  & 8.91 & 9.02 & 31.6 \\
         \hline
         2048  & 9.48 & 11.7 & 57.2 \\
         \hline
    \end{tabular}
    \label{tab:computeCost}
\end{table}

\subsection{Sparsity Characterization}
\label{subsec:batch_bridgeout_exp}
This section evaluates whether or not Batch Bridgeout, that is, a single set of Bridgeout perturbed weights per mini-batch, induce sparsity into the weights of convolutional layers. We trained the VGG-16 model without the fully connected layers. The detailed architecture of the model and the training method is described in Section~\ref{sec:vgg}. 

The network was trained with different regularization applied to all the convolutional layers. We use Hoyer's sparsity measure \cite{hoyer2004non} as a metric for quantifying sparsity. Hoyer's measure of a $d$-dimensional vector $x$ is given by
\begin{equation}
    H(\mathbf{x}) = \frac{\sqrt{d} - \frac{|\mathbf{x}|_1}{|\mathbf{x}|_2}}{\sqrt{d} - 1}
\end{equation}

\begin{figure*}
    \centering
    \includegraphics[width=0.95\linewidth]{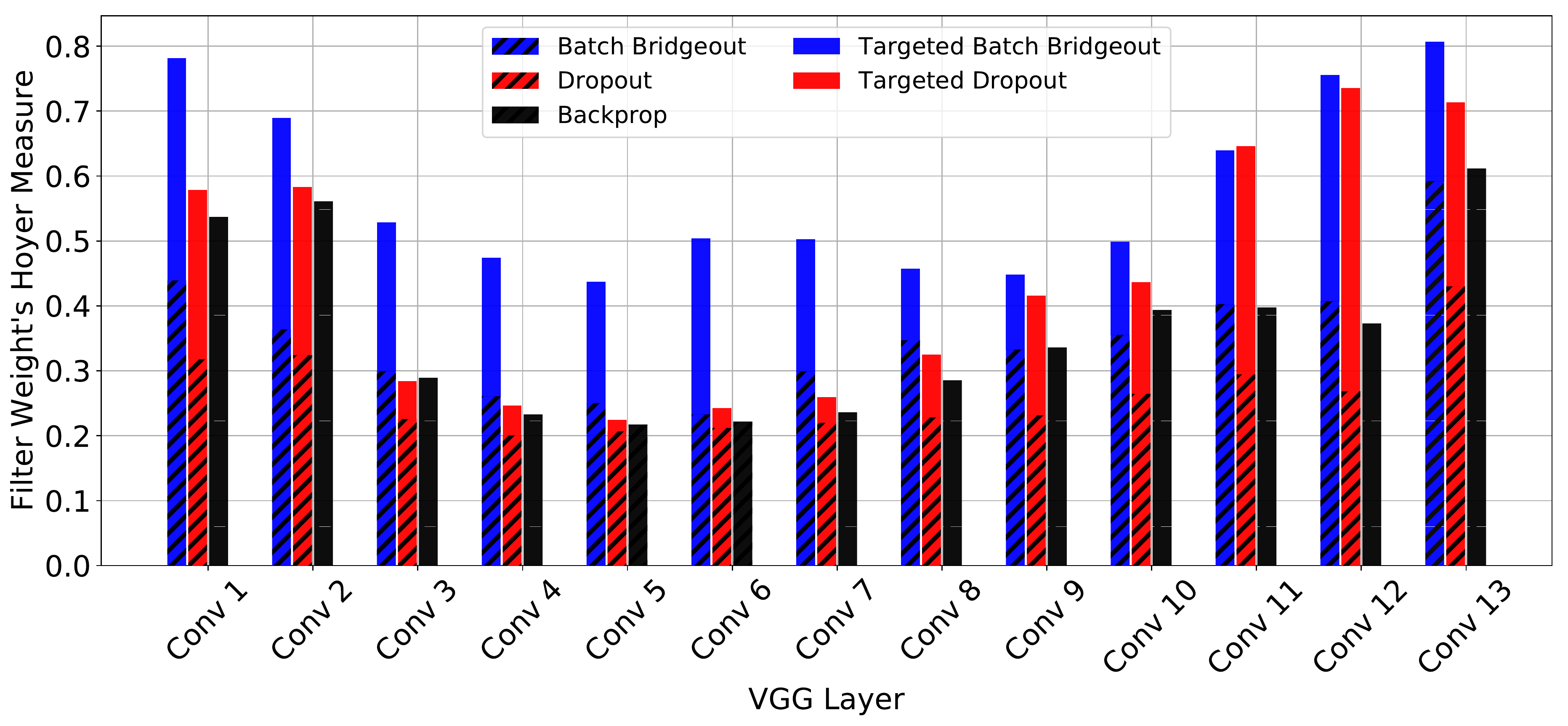}
    \caption{Sparsity of the weights of each convolutional layer of the VGG-16 architecture trained with different regularization techniques. The targeted application of Dropout is only slightly sparser than backpropagation alone. In almost all layers, targeted Batch Bridgeout results in the most sparse filters.}
    \label{fig:pruning_batch-bridgeout-weight-sparsity}
\end{figure*}
Figure~\ref{fig:pruning_batch-bridgeout-weight-sparsity} shows the Hoyer's sparsity measure of the filters of the convolution layers at the end of the training. As can be seen, Batch Bridgeout results in a higher sparsity measure compared to Dropout for almost all the layers. This shows that using Bridgeout with a single set of perturbed weights per mini-batch induces noticeable sparsity in the weights of the neural networks. We note that not applying any stochastic regularization results in higher sparsity for the layers near the input and output, this has been previously noted by Li et al. \cite{li2016pruning}. When the stochastic regularization is targeted only to the top 75\% of the weights, there is a significant increase in the sparsity of the weights. This motivates us to target Batch Bridgeout and Dropout only to the lowest magnitude weights for the purposes of pruning.

Targeting frees up the important weights from the effects of regularization and thus those important weights could grow larger, creating an imbalance in the distribution of weights. That is, regularized weights shrinking smaller and important weights being updated as dictated by the gradient of the cost function. This imbalance could result in higher sparsity of the targeted versions of the regularization. It can be seen that among all the methods, Targeted Batch Bridgeout results in the highest sparsity in the VGG architecture.

\begin{figure}
    \centering
    \includegraphics[width=0.9\linewidth]{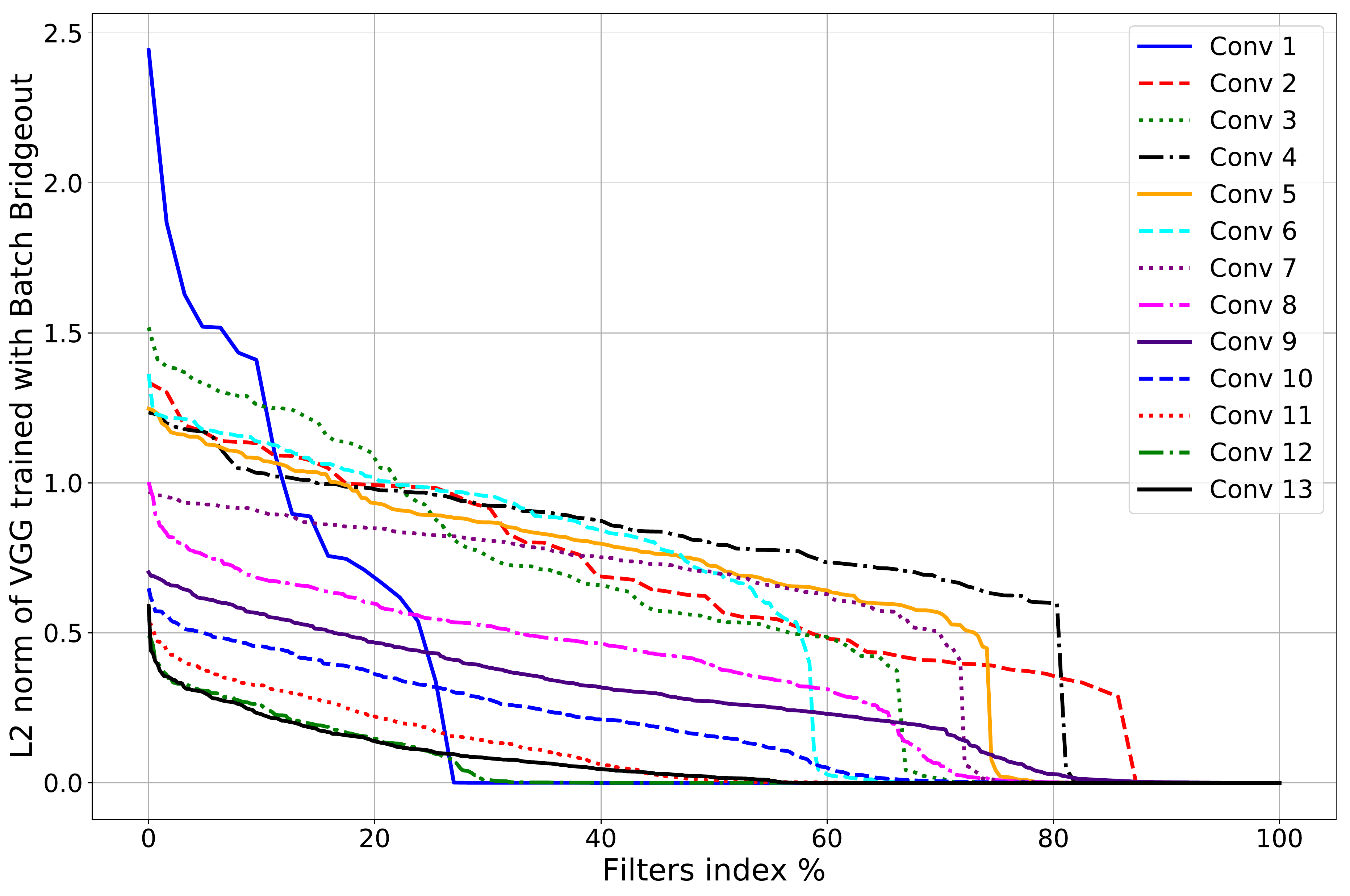}
    \includegraphics[width=0.9\linewidth]{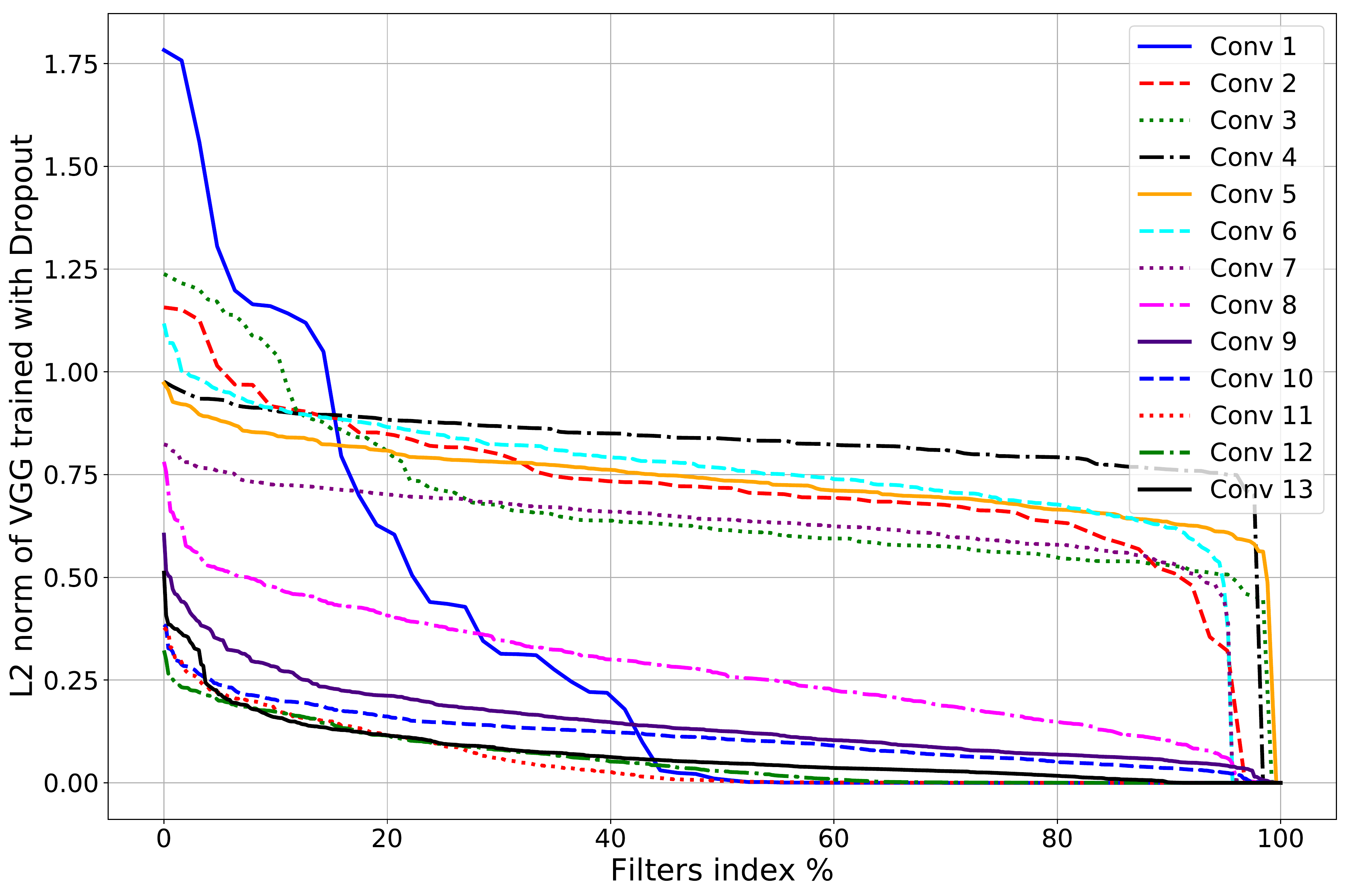}
    \caption{Sorted $L_2$ norm of the weights of the filters of VGG-16 trained with Batch Bridgeout (top) and Dropout (bottom). Targeted Batch Bridgeout results in steeper sloped $L_2$-norm filters, which is beneficial for pruning units/filters \cite{li2016pruning}. For Batch Bridgeout, the last 20\% of the filters in almost all layers is close to zero.}
    \label{fig:pruning_targeted-bridgeout-units-l2}
\end{figure}

Since we are using the $L_2$ norm as the importance criteria for keeping a filter during pruning, Figure~\ref{fig:pruning_targeted-bridgeout-units-l2} shows the slope of the $L_2$-norm of the filters each layer of VGG-16 trained with Batch Bridgeout and Dropout. As can be seen, Batch Bridgeout results in larger slopes of the filters compared to Dropout. It has been observed by Li et al. that layers with larger slopes maintain their accuracy as filters are pruned in that layer \cite{li2016pruning}.

\begin{figure}
    \centering
    \includegraphics[width=0.9\linewidth]{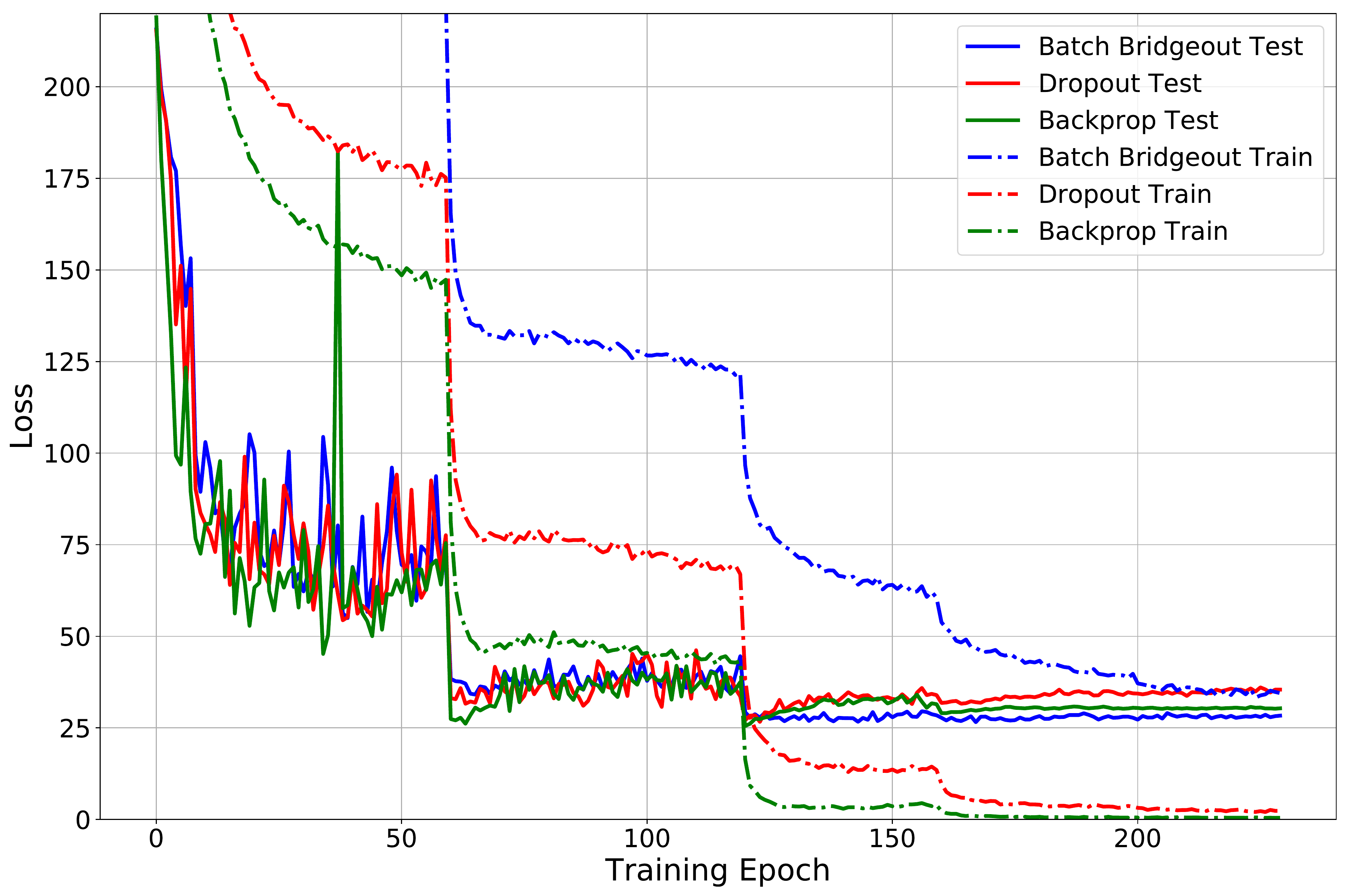}
    \includegraphics[width=0.9\linewidth]{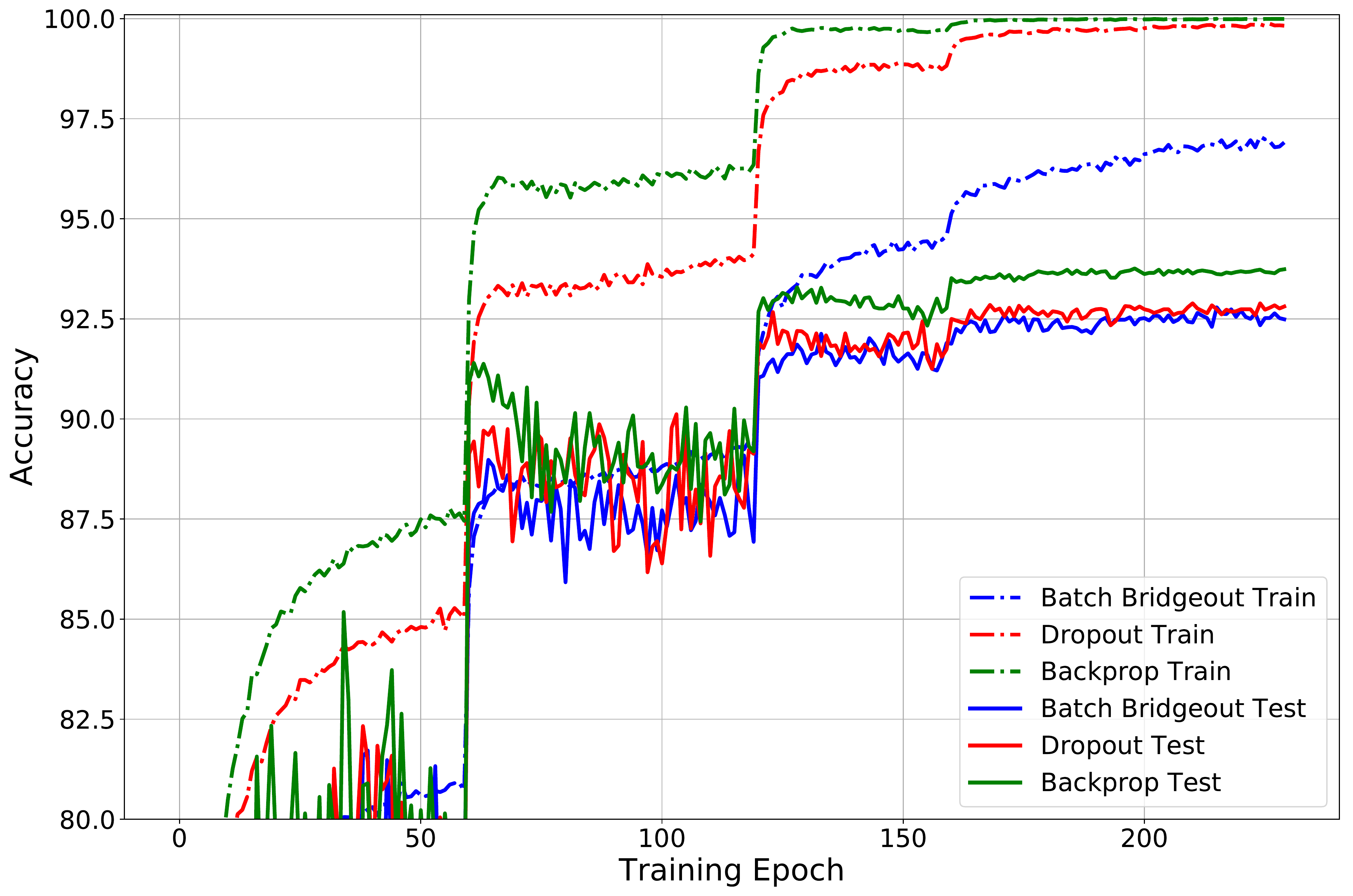}
    \caption{Training curves of VGG-16. Batch Bridgeout converges slightly later than backprop and dropout, this is typical of stochastic regularization techniques. The higher training loss of the Batch Bridgeout and Dropout is due to stochastic perturbations during training.}
    \label{fig:pruning_vgg-loss}
\end{figure}

Figure~\ref{fig:pruning_vgg-loss} shows the training and validation loss and accuracy for models trained with different techniques. As shown in the figure, all three models converge with Batch Bridgeout still having a downward trend whereas the validation loss starts ascending for the model without regularization. The training loss for Batch Bridgeout is higher compared to Dropout due to the large amount of perturbation applied during training.

\subsection{Pruning VGGNet}
\label{sec:vgg}
\begin{figure}
    \centering
    \includegraphics[width=0.9\linewidth]{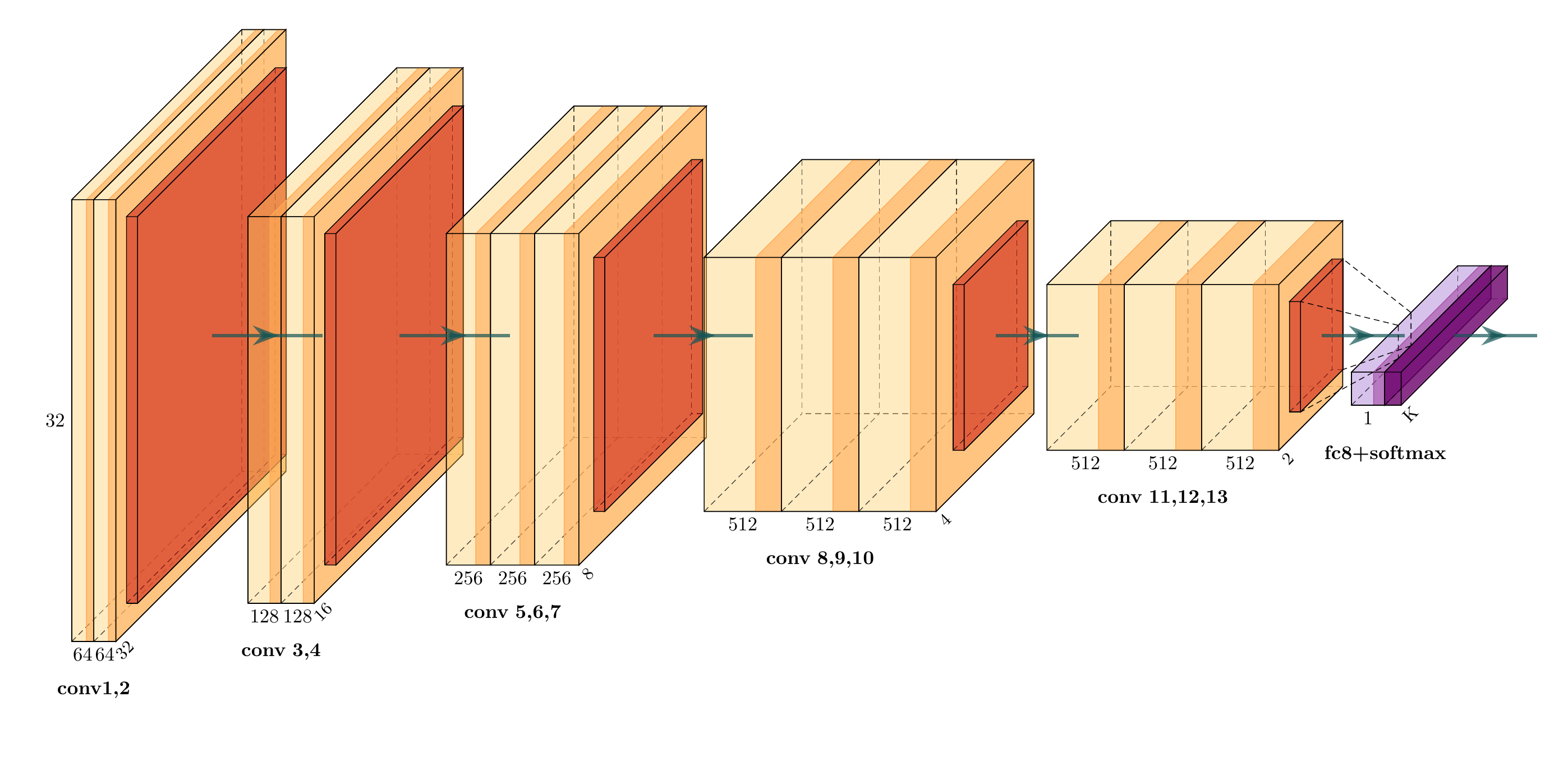}
    \caption{VGG-16 architecture without the fully connected layers used for evaluating the pruning performance of Batch Bridgeout.}
    \label{fig:pruning_vgg-arch}
\end{figure}

In this section we describe the pruning performance of Batch Bridgeout for the VGG architecture \cite{simonyan2014very}. The VGG architecture is one of the popular deep convolutional networks used in computer vision. The VGG-16 consists of 13 convolutional layers of receptive field of size $3\times3$ with some layers followed by max pooling, two fully connected layers of $4096$ units followed by softmax layer of 10 units representing the class probabilities. In order to make a more challenging pruning task, we train the VGG-16 without the two fully connected layers, which represent 90\% of the parameters of the VGG-16 model as shown in Figure.~\ref{fig:pruning_vgg-arch}. Starting with the small number of parameters, pruning any further parameters is difficult. Additionally, we add batch normalization~\cite{ioffe2015batch} after each convolutional layer.

Unless stated otherwise, all the networks were trained on the CIFAR-10~\cite{krizhevsky2009learning} image classification dataset for 230 epochs with stochastic gradient descent, an exponentially decaying learning rate of $0.1$, momentum of $0.9$ and weight decay of $5\times10^{-4}$. A Dropout probability of $p=0.3$ and Batch Bridgeout norm of $q=1.5$ was used. Both Batch Bridgeout and Dropout were targeted to the lowest magnitude $75\%$ of the weights in each layer except the last layer in the network.

\begin{figure}
    \centering
    \includegraphics[width=0.98\linewidth]{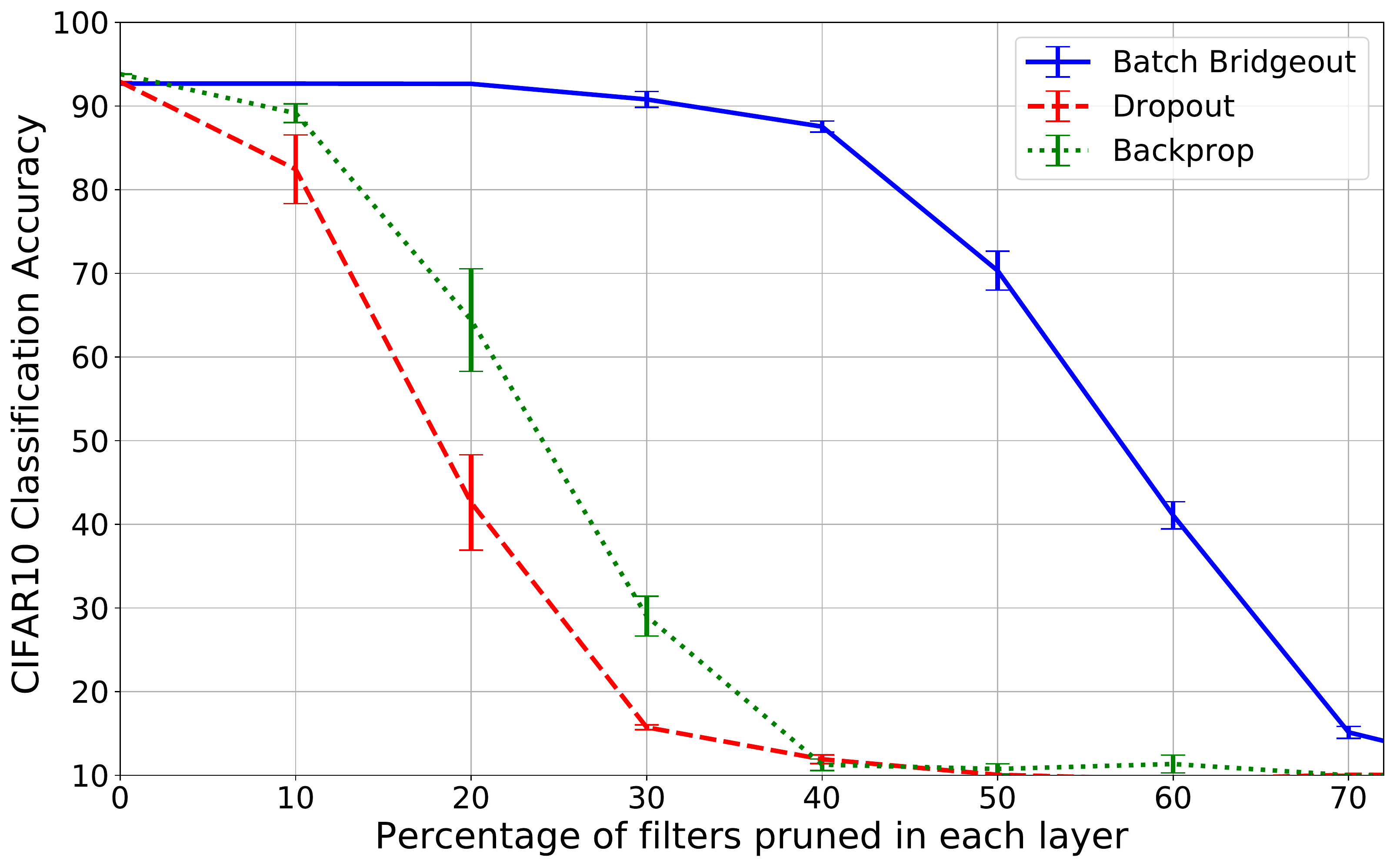}
    \caption{Accuracy of VGG-16 as a percentage of filters zeroed out, uniformly across all layers. Batch Bridgeout retains its original performance even when 40\% of the filters are removed from all the convolutional layers in the model.}
    \label{fig:pruning_CIFAR10-vggnet-unit-pruning}
\end{figure}

After training the networks with different regularization techniques, we zero out a constant fraction of low-magnitude filters across all convolutional layers, independently. Figure~\ref{fig:pruning_CIFAR10-vggnet-unit-pruning} shows the average accuracy of the VGG-16 model when different percentage of filters are set to zero. As can be seen, the degradation in accuracy of the Batch Bridgeout trained network is negligible when about $40\%$ of the filters in each layer are pruned. Whereas the Dropout and backprop trained networks degrades significantly when even $10\%$ of the filters are set to zero in the network.

In order to evaluate the computational cost of the pruned models (e.g., memory required by the model and the forward pass execution time) we remove the zeroed out filter rows from the weight matrices. Since removing a filter from the weight matrix in layer $l$ results in the reduction of the input size to the layer $l+1$, we remove all the weights in the $l+1$ layer corresponding to the removed filter's activation maps. For the final fully connected softmax layer, we remove the weights corresponding to the pruned out filters. Table~\ref{tab:vgg-pruned-cost} shows the memory and the average forward pass execution time at different levels of pruning and the accuracy of the models trained with different regularization. Similar to Figure~\ref{fig:pruning_CIFAR10-vggnet-unit-pruning}, in Table~\ref{tab:vgg-pruned-cost} Batch Bridgeout results in the least degradation as filters are removed. For the full model the backpropagation results in the highest accuracy indicating that the regularization techniques reduce the complexity of the model slightly in the absence of fully connected layers.

\begin{table*}[]
\centering
\caption{VGG-16 trained on CIFAR-10 dataset. Memory represents the footprint of the model in megabytes. Run-time is the execution time elapsed during inference of the CIFAR test set on 8-core CPU laptop. The last three columns show the test accuracy of the models trained with Batch Bridgeout, Dropout, and backpropagation.}
\vspace{0.1cm}
\begin{tabular}{|c|c|c|c||c|c|c|}
\hline
Filters pruned    & Memory       & Run-time   & Compression/ & Batch Bridgeout  & Dropout & Backprop \\
(\%)        & (MB)          & (Seconds) &   Speedup &  (\%) &  (\%) &   (\%)
\\ \hline\hline
0 & 56.18 &   32.53&  1/1        &92.79          &  92.89      & \textbf{93.76}\\ \hline
10 & 45.59 &   31.00& 1.23/1.05  &\textbf{92.79} &  90.87 & 87.40\\ \hline
20 & 36.05 &   25.68&  1.55/1.26 &\textbf{92.72} &  80.08 & 64.57\\ \hline
30 & 27.66 &   20.61&  2.03/1.58 &\textbf{90.93} &  55.67 & 48.14\\ \hline
40 & 20.35 &   16.40&  2.76/1.98 &\textbf{88.73} &  27.07 & 33.51\\ \hline
50 & 14.06 &   11.51&  4.00/2.82 &\textbf{76.14} &  14.00 & 19.79\\ \hline
60 & 9.04 &   9.30&   6.21/3.5 &\textbf{48.20}   &  10.42 & 13.27\\ \hline
70 & 5.10 &   6.85&   11.01/4.75&\textbf{24.38}  &  10.00 & 10.03\\ \hline
80 & 2.29 &   4.33&   24.5/7.51&10.00            &  10.18 & 10.00\\ \hline
90 & 0.58 &  2.31&   96.8/14.08 &10.00           &  10.00 & 10.00\\ \hline
\end{tabular}
\label{tab:vgg-pruned-cost}
\end{table*}

In order to determine which model is able to regain its original accuracy after pruning, we retrained the pruned models for $100$ epochs without any regularization. As a baseline, we also trained an equivalent sized model from random initialization.
Table~\ref{tab:vgg-pruned-retrained} shows the classification performance of the retrained models. For up to 40\% pruning the Batch Bridgeout is able to achieve its own original accuracy of $92.79$. 
The model trained from scratch comes close to the accuracy of the re-trained pruned models. This finding is consistent with previous work~\cite{li2016pruning} and calls into question the overall utility of post-training pruning vs. selecting a smaller size model \emph{a priori}. This is discussed in Section~\ref{sec:discuss} in the light of recent developments in the field.

\begin{table}[]
\centering
\caption{VGG-16 trained with different regularization techniques, pruned and retrained without any regularization to achieve the baseline accuracy. Scratch represents an equivalent sized VGG trained from random initialization.}
\begin{tabular}{|c|c|c|c|c|}
\hline
Pruning    & Bridgeout   & Dropout & Backprop & Scratch \\ \hline
20\% 		& 93.14  & \textbf{93.32} & 93.2 & 92.65\\ \hline
30\% 		& \textbf{93.38}  & 92.65 & 92.02 & 92.71\\ \hline
40\%		& \textbf{92.67}  & 92.03 & 91.29 & 92.22\\ \hline
50\% 		& \textbf{92.38}  & 90.93 & 90.64 & 91.84\\ \hline
\end{tabular}
\label{tab:vgg-pruned-retrained}
\end{table}

\subsection{Pruning ResNet and Wide-ResNet}
\label{subsec:resnet}
In this section we describe experiments to evaluate whether the pruning results obtained for VGG-16 generalize to small sized models and other state-of-the-art deep learning models. For this purpose, we evaluated a very small ResNet~\cite{he2016deep} model and a large Wide-ResNet-28x10~\cite{zagoruyko2016wide} on the CIFAR10 dataset. Unlike VGG, ResNet models include identity connections between alternating convolution layers in order to facilitate the training of very deep neural networks. In order to make for a more challenging pruning task, we selected a small all convolutional ResNet model with four residual blocks of $64,128,256$ and $512$ filters for a total of only $0.5$ million parameters. Since the model is already fairly small, removing any filters after training is expected to degrade task performance and thus makes for a good test for evaluating the pruning techniques. Wide-ResNet was selected as the standard implementation with $36$ million parameters.

\begin{figure}
\centering
\includegraphics[width=0.9\linewidth]{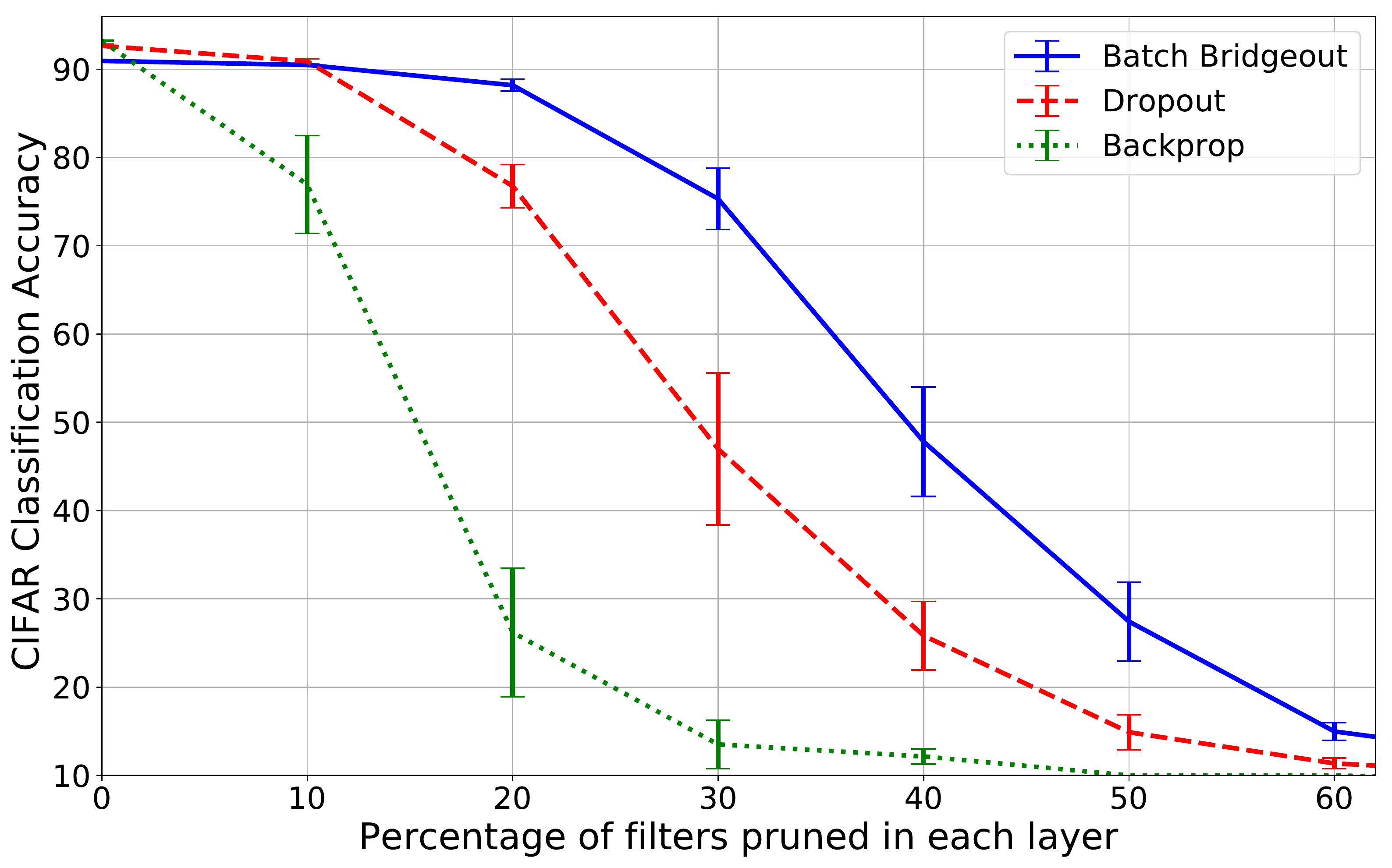}
\caption{Accuracy  of  ResNet  as  a  percentage  of  filters zeroed out, uniformly across all layers.}
\label{fig:pruning_cifar-resnet-unit-pruning}
\end{figure}

\begin{figure}
\centering
\includegraphics[width=0.9\linewidth]{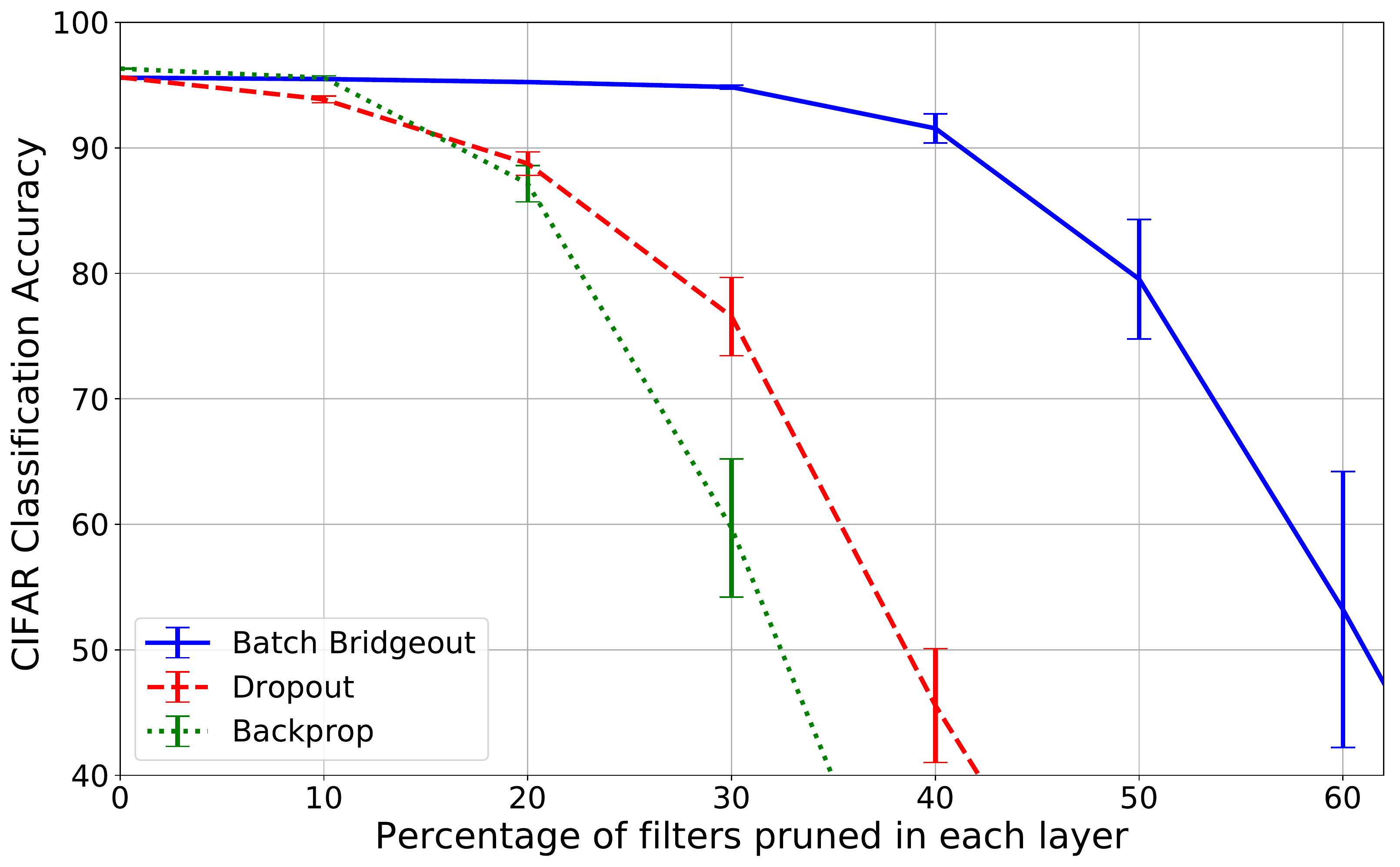}
\caption{Accuracy  of  Wide-ResNet-28x10  as  a  percentage of  filters zeroed out,  uniformly across all layers.}
\label{fig:pruning_cifar-wide-resnet-unit-pruning}
\end{figure}

Figure~\ref{fig:pruning_cifar-resnet-unit-pruning} and Figure~\ref{fig:pruning_cifar-wide-resnet-unit-pruning} show the classification accuracy of ResNet and Wide-ResNet as a percentage of filters are zeroed out in each layer uniformly, respectively. In both the networks, we see the same relative results where Batch Bridgeout achieved the highest accuracy compared to Dropout and backpropagation.


\section{Discussion}
\label{sec:discuss}

Higher performance on benchmark computer vision tasks has been achieved by increasingly larger and deeper convolutional neural networks, such as VGGNet~\cite{simonyan2014very}, ResNet~\cite{he2016deep} and Wide-ResNet~\cite{zagoruyko2016wide}.
To deploy these models to resource limited devices post-training model pruning is often required. For this reason, model pruning is an active area of research in deep learning. Structured-pruning, where entire filters are removed, provides the best resource savings because removing filters reduces the dimension of the matrix multiplication. However, structured pruning results in more severe performance degradation compared to unstructured pruning of individual weights. We expect that regularization techniques that promote sparsity and result in robust networks will be useful in avoiding degradation in performance due to structured pruning. Regularization with Batch Bridgeout, proposed in this paper, provides both sparsity and stochasticity for robustness and is computationally efficient for use with large models.
Our experiments demonstrated consistent improvement in pruned-model accuracy for Batch Bridgeout compared to Dropout or backpropagation with weight decay across a range of deep learning architectures.

Our results for retraining pruned models demonstrated that Batch Bridgeout allowed models to better regain their original accuracy, even for 40\% pruning on the fully-convolutional version of VGG16. However, we also observed that training a similarly sized model from scratch was typically within 1\% of the retrained pruned model for CIFAR-10. This gives rise to the question of whether it is better to train a smaller model rather than training a large model and then performing post-training pruning. 

Recently, several studies have tried to investigate this question. Frankle and Carbin~\cite{frankle2019the} performed experiments with LeNet-5~\cite{lecun1990handwritten} to show that a sub-network obtained from pruning the original larger model could be trained to the same accuracy as the original model \textit{if the weights of the sub-network are initialized with its initial random weights} when it was part of the original model. They argue that training and pruning helps in discovering this special sub-network which could perform as good as the original model. Liu et al.~\cite{liu2018rethinking} experimentally showed that pruned sub-networks could be initialized with \textit{random weights} and trained to achieve the performance of the original model. Gale et al.~\cite{gale2019state} performed large scale computer vision and neural machine translation experiments and concluded that for complex tasks unstructured pruned models cannot be trained to the same task performance from scratch as could be obtained from optimizing and pruning.
Gale et al. investigated unstructured pruning which could be seen as an upper bound on the performance of an equivalent sized structurally pruned model. That is, an unstructured pruned model still inherits the surviving-weights' topology, whereas structured pruned models initialized from scratch do not inherit any information from the pruning process. Thus, given the results of Gale et al., it is unlikely that for complex tasks small models equivalent to a structurally pruned model will attain the same performance as the pruned models. Furthermore, given the fact that Batch Bridgeout is able to retain performance when pruning very small models indicates the utility of pruning even when small models are used for relatively simple computer vision tasks.

One reason for the good performance of the pruned models with random initialization could be the long-term use of the standard benchmark datasets such as CIFAR for building computer vision models. As noted by Recht et al.~\cite{recht2018cifar}, sampling a different test set from the CIFAR-10 dataset results in large changes in the generalization error of contemporary computer vision models. A plethora of hyperparameters and architecture designs have been tailored towards these benchmark datasets. This could mean that under these settings, the number of trainable hyperparameters or model size could be reduced quite a bit without sacrificing the performance on the given test set. However, for solving novel tasks large models might still be needed. As future work, we plan to investigate this question with large scale complex tasks such as evaluation on Imagenet~\cite{deng2009imagenet}.

Nevertheless, our results indicate that, under identical conditions, Batch Bridgeout trained networks are more robust to structured pruning compared to Dropout and backpropagation across several architectures and models of different sizes. These results also hold when the architectures are thinned out in the first place, e.g. by removing 90\% of the parameters of VGG16 by omitting the two large fully connected layers, and by selecting a small ResNet model with less than 0.5M parameters.

In the present study, all layers in a DNN were pruned to the same degree. However, in a DNN, some layers are more sensitive to pruning verses others as shown in Figure.~\ref{fig:pruning_batch-bridgeout-weight-sparsity}. As future work, the amount of pruning performed could be adjusted on a layer-by-layer basis, which could potentially result in better accuracy for a given amount of pruning.

\section{Summary}
\label{sec:summary}
 In this paper, we show that effective sparsity inducing regularization techniques are important for compressing large neural network models. We presented Batch Bridgeout, a computationally efficient extension of the Bridgeout regularization scheme, and demonstrated empirically that it is capable of inducing sparsity in the filters of convolutional neural networks. Batch Bridgeout was evaluated on structured filter pruning on a number of CNN architectures, including VGG, ResNet and Wide-ResNet. For all architectures, Batch Bridgeout was shown to outperform the recently proposed Targeted Dropout and weight decay regularization based pruning. 
 
\begin{ack}
This research was undertaken thanks in part to funding from the Canada First Research Excellence Fund and the Natural Sciences and Engineering Research Council of Canada.
\end{ack}
\bibliographystyle{apalike}
\bibliography{egbib}

\begin{thebibliography}{}

\bibitem[Aich et~al., 2020]{aich2020multi}
Aich, S., Yamazaki, M., Taniguchi, Y., and Stavness, I. (2020).
\newblock Multi-scale weight sharing network for image recognition.
\newblock {\em Pattern Recognition Letters}, 131:348--354.

\bibitem[Alvarez and Salzmann, 2016]{alvarez2016learning}
Alvarez, J.~M. and Salzmann, M. (2016).
\newblock Learning the number of neurons in deep networks.
\newblock In {\em Advances in Neural Information Processing Systems}, pages
  2270--2278.

\bibitem[Bui et~al., 2019]{bui2019ell_0}
Bui, K., Park, F., Zhang, S., Qi, Y., and Xin, J. (2019).
\newblock $l_0$ regularized structured sparsity convolutional neural networks.
\newblock {\em arXiv preprint arXiv:1912.07868}.

\bibitem[Chetlur et~al., 2014]{chetlur2014cudnn}
Chetlur, S., Woolley, C., Vandermersch, P., Cohen, J., Tran, J., Catanzaro, B.,
  and Shelhamer, E. (2014).
\newblock cudnn: Efficient primitives for deep learning.
\newblock {\em arXiv preprint arXiv:1410.0759}.

\bibitem[Deng et~al., 2009]{deng2009imagenet}
Deng, J., Dong, W., Socher, R., Li, L.-J., Li, K., and Fei-Fei, L. (2009).
\newblock Imagenet: A large-scale hierarchical image database.
\newblock In {\em 2009 IEEE conference on computer vision and pattern
  recognition}, pages 248--255. Ieee.

\bibitem[Frankle and Carbin, 2019]{frankle2019the}
Frankle, J. and Carbin, M. (2019).
\newblock The lottery ticket hypothesis: Finding sparse, trainable neural
  networks.
\newblock In {\em International Conference on Learning Representations}.

\bibitem[Gale et~al., 2019]{gale2019state}
Gale, T., Elsen, E., and Hooker, S. (2019).
\newblock The state of sparsity in deep neural networks.
\newblock {\em arXiv preprint arXiv:1902.09574}.

\bibitem[Gale et~al., 2020]{gale2020sparse}
Gale, T., Zaharia, M., Young, C., and Elsen, E. (2020).
\newblock Sparse gpu kernels for deep learning.
\newblock {\em arXiv preprint arXiv:2006.10901}.

\bibitem[Glorot et~al., 2011]{glorot2011deep}
Glorot, X., Bordes, A., and Bengio, Y. (2011).
\newblock Deep sparse rectifier neural networks.
\newblock In {\em Proceedings of the Fourteenth International Conference on
  Artificial Intelligence and Statistics}, pages 315--323.

\bibitem[Gomez et~al., 2019]{gomez2019learning}
Gomez, A.~N., Zhang, I., Swersky, K., Gal, Y., and Hinton, G.~E. (2019).
\newblock Learning sparse networks using targeted dropout.
\newblock {\em arXiv preprint arXiv:1905.13678}.

\bibitem[Gysel et~al., 2018]{gysel2018ristretto}
Gysel, P., Pimentel, J., Motamedi, M., and Ghiasi, S. (2018).
\newblock Ristretto: A framework for empirical study of resource-efficient
  inference in convolutional neural networks.
\newblock {\em IEEE transactions on neural networks and learning systems},
  29(11):5784--5789.

\bibitem[Han et~al., 2015]{han2015learning}
Han, S., Pool, J., Tran, J., and Dally, W. (2015).
\newblock Learning both weights and connections for efficient neural network.
\newblock In {\em Advances in Neural Information Processing Systems}, pages
  1135--1143.

\bibitem[He et~al., 2016]{he2016deep}
He, K., Zhang, X., Ren, S., and Sun, J. (2016).
\newblock Deep residual learning for image recognition.
\newblock In {\em Proceedings of the IEEE conference on computer vision and
  pattern recognition}, pages 770--778.

\bibitem[Hestness et~al., 2017]{hestness2017deep}
Hestness, J., Narang, S., Ardalani, N., Diamos, G., Jun, H., Kianinejad, H.,
  Patwary, M., Ali, M., Yang, Y., and Zhou, Y. (2017).
\newblock Deep learning scaling is predictable, empirically.
\newblock {\em arXiv preprint arXiv:1712.00409}.

\bibitem[Hinton et~al., 2015]{hinton2015distilling}
Hinton, G., Vinyals, O., and Dean, J. (2015).
\newblock Distilling the knowledge in a neural network.
\newblock {\em arXiv preprint arXiv:1503.02531}.

\bibitem[Hoyer, 2004]{hoyer2004non}
Hoyer, P.~O. (2004).
\newblock Non-negative matrix factorization with sparseness constraints.
\newblock {\em Journal of machine learning research}, 5(Nov):1457--1469.

\bibitem[Huttenlocher et~al., 1979]{huttenlocher1979synaptic}
Huttenlocher, P.~R. et~al. (1979).
\newblock Synaptic density in human frontal cortex-developmental changes and
  effects of aging.
\newblock {\em Brain Res}, 163(2):195--205.

\bibitem[Inan et~al., 2016]{inan2016tying}
Inan, H., Khosravi, K., and Socher, R. (2016).
\newblock Tying word vectors and word classifiers: A loss framework for
  language modeling.
\newblock {\em arXiv preprint arXiv:1611.01462}.

\bibitem[Ioffe and Szegedy, 2015]{ioffe2015batch}
Ioffe, S. and Szegedy, C. (2015).
\newblock Batch normalization: Accelerating deep network training by reducing
  internal covariate shift.
\newblock In {\em International Conference on Machine Learning}, pages
  448--456.

\bibitem[Khan et~al., 2018]{khan2018bridgeout}
Khan, N., Shah, J., and Stavness, I. (2018).
\newblock Bridgeout: Stochastic bridge regularization for deep neural networks.
\newblock {\em IEEE Access}, 6:42961--42970.

\bibitem[Krizhevsky and Hinton, 2009]{krizhevsky2009learning}
Krizhevsky, A. and Hinton, G. (2009).
\newblock Learning multiple layers of features from tiny images.
\newblock Technical report, Citeseer.

\bibitem[LeCun et~al., 1990]{lecun1990handwritten}
LeCun, Y., Boser, B.~E., Denker, J.~S., Henderson, D., Howard, R.~E., Hubbard,
  W.~E., and Jackel, L.~D. (1990).
\newblock Handwritten digit recognition with a back-propagation network.
\newblock In {\em Advances in neural information processing systems}, pages
  396--404.

\bibitem[LeCun et~al., 1989]{lecun1989optimal}
LeCun, Y., Denker, J.~S., Solla, S.~A., Howard, R.~E., and Jackel, L.~D.
  (1989).
\newblock Optimal brain damage.
\newblock In {\em NIPS}, volume~2, pages 598--605.

\bibitem[Li et~al., 2016]{li2016pruning}
Li, H., Kadav, A., Durdanovic, I., Samet, H., and Graf, H.~P. (2016).
\newblock Pruning filters for efficient convnets.
\newblock {\em arXiv preprint arXiv:1608.08710}.

\bibitem[Liu et~al., 2018]{liu2018rethinking}
Liu, Z., Sun, M., Zhou, T., Huang, G., and Darrell, T. (2018).
\newblock Rethinking the value of network pruning.
\newblock In {\em International Conference on Learning Representations}.

\bibitem[Louizos et~al., 2018]{louizos2018learning}
Louizos, C., Welling, M., and Kingma, D.~P. (2018).
\newblock Learning sparse neural networks through $l_0$ regularization.
\newblock In {\em International Conference on Learning Representations}.

\bibitem[Molchanov et~al., 2017]{molchanov2017variational}
Molchanov, D., Ashukha, A., and Vetrov, D. (2017).
\newblock Variational dropout sparsifies deep neural networks.
\newblock In {\em Proceedings of the 34th International Conference on Machine
  Learning-Volume 70}, pages 2498--2507. JMLR. org.

\bibitem[Novikov et~al., 2015]{novikov2015tensorizing}
Novikov, A., Podoprikhin, D., Osokin, A., and Vetrov, D.~P. (2015).
\newblock Tensorizing neural networks.
\newblock In {\em Advances in neural information processing systems}, pages
  442--450.

\bibitem[Recht et~al., 2018]{recht2018cifar}
Recht, B., Roelofs, R., Schmidt, L., and Shankar, V. (2018).
\newblock Do cifar-10 classifiers generalize to cifar-10?
\newblock {\em arXiv preprint arXiv:1806.00451}.

\bibitem[Safran and Shamir, 2016]{Safran2016}
Safran, I. and Shamir, O. (2016).
\newblock On the quality of the initial basin in overspecified neural networks.
\newblock In {\em Proceedings of the 33rd International Conference on
  International Conference on Machine Learning - Volume 48}, ICML’16, page
  774–782. JMLR.org.

\bibitem[Scardapane et~al., 2017]{scardapane2017group}
Scardapane, S., Comminiello, D., Hussain, A., and Uncini, A. (2017).
\newblock Group sparse regularization for deep neural networks.
\newblock {\em Neurocomputing}, 241:81--89.

\bibitem[Shazeer et~al., 2017]{ShazeerMMDLHD17}
Shazeer, N., Mirhoseini, A., Maziarz, K., Davis, A., Le, Q.~V., Hinton, G.~E.,
  and Dean, J. (2017).
\newblock Outrageously large neural networks: The sparsely-gated
  mixture-of-experts layer.
\newblock In {\em ICLR (Poster)}. OpenReview.net.

\bibitem[Simonyan and Zisserman, 2014]{simonyan2014very}
Simonyan, K. and Zisserman, A. (2014).
\newblock Very deep convolutional networks for large-scale image recognition.
\newblock {\em arXiv preprint arXiv:1409.1556}.

\bibitem[Srivastava et~al., 2014]{srivastava2014dropout}
Srivastava, N., Hinton, G.~E., Krizhevsky, A., Sutskever, I., and
  Salakhutdinov, R. (2014).
\newblock Dropout: a simple way to prevent neural networks from overfitting.
\newblock {\em Journal of Machine Learning Research}, 15(1):1929--1958.

\bibitem[Ullrich et~al., 2017]{ullrich2017soft}
Ullrich, K., Meeds, E., and Welling, M. (2017).
\newblock Soft weight-sharing for neural network compression.
\newblock {\em arXiv preprint arXiv:1702.04008}.

\bibitem[Wen et~al., 2017]{wen2017learning}
Wen, W., He, Y., Rajbhandari, S., Zhang, M., Wang, W., Liu, F., Hu, B., Chen,
  Y., and Li, H. (2017).
\newblock Learning intrinsic sparse structures within long short-term memory.
\newblock {\em arXiv preprint arXiv:1709.05027}.

\bibitem[Xie et~al., 2019]{xie2019self}
Xie, Q., Hovy, E., Luong, M.-T., and Le, Q.~V. (2019).
\newblock Self-training with noisy student improves imagenet classification.
\newblock {\em arXiv preprint arXiv:1911.04252}.

\bibitem[Yoon and Hwang, 2017]{yoon2017combined}
Yoon, J. and Hwang, S.~J. (2017).
\newblock Combined group and exclusive sparsity for deep neural networks.
\newblock In {\em Proceedings of the 34th International Conference on Machine
  Learning-Volume 70}, pages 3958--3966. JMLR. org.

\bibitem[Zagoruyko and Komodakis, 2016]{zagoruyko2016wide}
Zagoruyko, S. and Komodakis, N. (2016).
\newblock Wide residual networks.
\newblock {\em arXiv preprint arXiv:1605.07146}.

\end{thebibliography}

\end{document}